\documentclass{article}
\usepackage{ijcai22}

\usepackage{times}
\usepackage{soul}
\usepackage{url}
\usepackage[hidelinks]{hyperref}
\usepackage[utf8]{inputenc}
\usepackage[small]{caption}
\usepackage{graphicx}
\usepackage{amsmath}
\usepackage{amsthm}
\usepackage{booktabs}
\usepackage{algorithm}
\usepackage{algorithmic}
\title{Using calibrator to improve robustness in Machine Reading Comprehension}

\author{
Jing Jin
\and
Houfeng Wang
\affiliations
Key Laboratory of Computational Linguistics, Peking University, MOE, China
\emails
\{11jj617, wanghf\}@pku.edu.cn
}

\begin{document}

\maketitle

\begin{abstract}
 Machine Reading Comprehension(MRC) has achieved a remarkable result since some powerful models, such as BERT, are proposed. However, these models are not robust enough and vulnerable to adversarial input perturbation and generalization examples. Some works tried to improve the performance on specific types of data by adding some related examples into training data while it leads to degradation on the original dataset, because the shift of data distribution makes the answer ranking based on the softmax probability of model unreliable. In this paper, we propose a method to improve the robustness by using a calibrator as the post-hoc reranker, which is implemented based on XGBoost model. The calibrator combines both manual features and representation learning features to rerank candidate results. Experimental results on adversarial datasets show that our model can achieve performance improvement by more than 10\% and also make improvement on the original and generalization datasets.

\end{abstract}

\section{Introduction}
Assisted by large pre-trained models, Machine Reading Comprehension(MRC) has achieved human-comparable results on some existing datasets. But even state-of-the-art(SOTA) models trained on such datasets are not robust enough. These models are not only vulnerable to adversarial input perturbations, but also perform poorly on out-of-domain data. 

Building new MRC datasets that are more challenging and cover more domains may improve the robustness, but the whole process is expensive in both time and cost. Therefore, there are two ways to address the problem based on existing data. One is the data level. Using some of adversarial examples or out-of-domain examples as data augmentation can improve performance on corresponding dataset, but it leads to degradation on the original dataset. It is impractical to sacrifice the original prediction accuracy for the purpose of defense. The other is the model level. Adding complex structures in MRC models can slightly improve robustness and defend adversarial attack, but the new pipeline is time-consuming and memory intensive during training and inference.

In this paper, we proposed a simple yet effective method to improve performance on adversarial and generalization datasets without sacrificing original performance in extractive MRC task.
We implemented several general kinds of adversarial examples generation to explore the vulnerability of SOTA model. And we found that performance degradation is not because the model has completely lost its ability to predict the correct range of answers, but for the ranking of candidate answers becomes unreliable. In other words, the model can still predict the correct range, but won't choose it as final output. Based on the above observation and inspired by previous work, we proposed a MRC method, in which a calibrator is used as the post-hoc reranker to adjust the ranking of candidates. On account of the time complexity and space consumption for practical potential in realistic scenario, we adopted XGBoost to implement the calibrator. 

Instead of BERT~\cite{devlin2018bert}, we used RoBERTa~\cite{liu2019roberta} as our backbone MRC model, for the latter uses more data and refines dynamic masking strategy and shows higher level of robustness on MRC task . We use SQuAD 2.0 dataset~\cite{rajpurkar2018know} as main dataset and Natural Questions~\cite{kwiatkowski2019natural} as a representative of generalization datasets. We employ the methods proposed by ~\cite{maharana2020adversarial} to generate adversarial examples, which has been proved aggressive to attack baseline MRC models. And then we utilized our proposed calibrator as a post-hoc reranker to get better results on generated datasets and original datasets. 

Our contributions can be summarized as follows:
\begin{itemize}
\item We had a thorough research on adversarial examples generation on MRC datasets and made an analysis with statistical data of the influence of adversarial examples to MRC models.
\item We proposed a simple yet effective method to use calibrator as a reranker to improve performance on adversarial datasets without sacrificing original performance.
\item Our calibrator only takes less than ten minutes to train but shows noteworthy improvement. So the method we proposed is time-friendly and space-saving comparing to previous work that apply complex structures and have huge parameters. 
\end{itemize}

\section{Related Work}

\paragraph{Robustness in MRC} Robustness is a research highlight in NLP because researchers have found that models achieved impressive performance on particular datasets is too vulnerable for practical application~\cite{jin2020bert}. As for MRC, the research on robustness of models can be generally categorised into two directions: generalization to out-of-domain distributions and robustness under test-time perturbations~\cite{si2020benchmarking}. Both directions will disturb the data distribution, but they are different in goals. Adversarial input perturbations aim to ascertain whether model learns shortcut, that is to say model learns to answer questions based on some specific implicit sentence patterns rather than reading comprehension ability~\cite{lai2021machine}. Generalization aims to extend application scope of the model to out-of-domain data and maintains great performance under domain-shift~\cite{kamath2020selective}. Many previous researches have focus on exposing models' vulnerabilities through maliciously designed inputs and bringing forward to new challenging datasets and tools for training and evaluating~\cite{sen2020models,jin2020bert,si2020benchmarking,bartolo2021improving,si2021better}. Another perspective is to improve the robustness of model by modifying the model structure, such as introducing external knowledge and multi-task strategy~\cite{wu2020improvingrobustness}, adding adapters~\cite{han2021robust}, changing loss function to adjust bias caused by generalization~\cite{wu2020improving} and so on.  These models are more robust than the original but have more than doubled parameters. 

\paragraph{Adversarial Examples Generation Methods} The goal of adversarial attack is to mislead the model into giving wrong outputs. Due to discrete characteristics of Natural Language, some aggressive adversarial attack methods in Computer Vision may cause out-of-distribution(OOD) problem in NLP. In the domain of MRC, adversarial input perturbation on contexts and questions may have a great effect. There are several kinds of ways to perturb the text of contexts and questions, such as search, heuristics or gradient-based techniques and so on~\cite{zhang2019generating}. ~\cite{jia2017adversarial} first proposed to use QA examples with distracting sentences that have significant overlap with the question and are randomly inserted into the context as adversarial examples. However, the creation of such distracting sentences is based on some fixed templates, so the model probably identifies learnable biases and overfits to the templates instead of being robust to the attack itself~\cite{maharana2020adversarial}. Then more researches have tried to address this problem by creating more complex templates~\cite{wang2018robust}, using more dynamic strategies e.g. word substitution~\cite{bao2021defending}, or exploring more challenging generative methods~\cite{si2021better,bartolo2021improving}. In addition to adding confusing sentences into contexts, there are several methods that can be aggressive and cause huge performance degradation as well, such as deleting pivotal sentences from contexts~\cite{maharana2020adversarial}, using language models to generate new questions with same semantics and different syntactic forms as the origin~\cite{iyyer2018adversarial} and perturbing word embedding~\cite{lee2021learning}. 

\paragraph{Calibration in NLP} The question of whether a model's posterior probabilites provide an accurate empirical measure of how likely the model is to be correct on a given example has been put forward to examine the reliability of the model~\cite{desai2020calibration}. A well-calibrated model should have output probabilities align with the true probabilities of its predictions, which means the confidences of their predictions should be more aligned with their accuracies. Previous works have found that the model which gives good confidence estimates on in-domain data is overconfident on OOD data~\cite{desai2020calibration,kamath2020selective}. In MRC, models tend to choose results with max softmax probabilites as final outputs. But out-of-domain data leads to the shift of data distribution, so the softmax probabilies are unreliable estimates of confidence~\cite{kamath2020selective,xin2021art}. Previous works used to apply the calibrator as a threshold to decide whether to abstain the prediction and try to avoid making confident yet incorrect predictions on the preserved examples ~\cite{kamath2020selective,xin2021art}. Instead of using it as a threshold, we use the calibrator as a reranker based on the inspiration of ~\cite{zhang2021knowing} and our analysis on bad cases in MRC task. 

\section{Method}\label{section::3}
We use the calibrator as a post-hoc reranker to improve robustness of SOTA models in extractive MRC task. Basic QA model feeds outputs and some important model features into the calibrator and then calibrator chooses the best answer span from $k$ candidates as final outputs.
We follow prior works~\cite{kamath2020selective,zhang2021knowing}, for the idea and feature space. But we adopt different calibrator architecture and improve its feature space. We categorized features into two kinds: manual features that are irrelevant to the MRC model, and representation learning features that revealed model states.
\subsection{Metrics}
Previous works~\cite{kamath2020selective,zhang2021knowing} use a calibrator to decide whether to abstain an example, so the metrics to evaluate calibrator performance are associated with accuracy of binary classification and MRC performance of the retained examples. They plot risk versus coverage graph where coverage is the fraction of evaluation data that calibration chooses to retain and risk is the error at that coverage. And they calculate the area under the curve, i.e., AUROC( Area Under the Receiver Operating Characteristics Curve), as the metrics. In addition, they pay special attention to coverages when the accuracy is 80\% and 90\% and use them as metrics as well. A good calibrator should cover as much coverage as it can under a particular given accuracy.

We propose to use the calibrator as a reranker to choose the best, so it is a multi-classification problem rather than binary classification. And we don't abstain examples, so we use different metrics to evaluate the performance of calibrator. Classification accuracy is used to evaluate our calibrator.

To measure MRC task performance, we use the answer chosen by our calibrator as final outputs, and measure the answer span exact match(EM) score and F1 score like common extractive MRC task. We compare the EM score and F1 score with and without calibrator in two settings(trained on clean data or mixed data) to prove the effect of calibrator.

\subsection{Basic MRC model}
We use standard span prediction architecture, and choose RoBERTa-large~\cite{liu2019roberta} as our backbone model for its predominant performance and relatively low training cost. 

We don't modify the architecture of MRC model. The model has same input format and training process with general MRC models. But we make minor changes to the format of model's outputs. The model outputs the example's unique id and text of answer with max softmax probability as usual, it also needs to output top $k$ answers with their texts, start and end logits, and softmax probability, where $k$ is 10 in our experiments. And to train the calibrator, we label the best answer among top $k$ answers through the F1-score calculation.

And the hidden states of model are also included in the output if batch size is set to 1 to characterize model specific states under particular examples.

\subsection{Calibrator architecture}
A multi-classifier is trained using the gradient boosting library XGBoost~\cite{chen2016xgboost}, which chooses one answer from $k$ candidates provided by the baseline MRC model. The calibrator architecture does not share its weights with basic MRC models. Since our target is to prove the effect of calibrator on adversarial datasets, we simply keep most of hyperparameters as their default values: max depth, subsample, colsample by tree and so on. To accelerate the training and inference process, we set the number of estimators to 160 and set the learning rate to 0.1. There may be some space for improvement by tuning these hyperparameters, but we focus on the overall effect of calibrator on adversarial data, so there is no experiment related to tuning the hyperparameters.

\subsection{Manual features}
As said before, manual features are completely irrelevant to the model, but characterized the property of the data.

We use the following features for input example $i$: $q_i$ and $c_i$ indicate the text length of corresponding question and context respectively, $K_i$ is the collection of its $k$ candidates. For each candidate $k_{ij}$ in $K_i$ where $j$ is its original ranking in the candidates, we denote its features with a quadruple: ($l_{ij}$,$p_{ij}$,$s_{ij}$,$e_{ij}$), where $l_{ij}$ means the text length of the $j$-th candidate prediction of example $i$, $p_{ij}$ indicates corresponding softmax probability, $s_{ij}$ and $e_{ij}$ refer to start logits and end logits respectively.

Inspired by ~\cite{zhang2021knowing}, we proposed two heuristic features based on a small amount of additional calculation on the above features. 

One is based on the softmax probability of top $k$ predictions and calculates the entropy to integrate the entire candidate predictions. According to general calculation formula of entropy, the entropy feature $E_i$ we designed is calculated as:

\begin{equation}
    \resizebox{.91\linewidth}{!}{$
    \displaystyle
    E_i = -\left[\sum_{j=1}^k p_{ij}\log{p_{ij}}+\left(1-\sum_{j=1}^k p_{ij}\right)\log{\left(1-\sum_{j=1}^k p_{ij}\right)} \right]
$}
\end{equation}

The other is based on the calculation of softmax probability. When calculating the softmax probability for each candidate prediction, start and end logits are added as final score. And then we take the difference between each final score and the maximum value of all final scores to calculate the softmax probability. But the shift of data distribution leads to overconfident problem, so we use a scaling factor $\lambda$ to alleviate the problem. The whole calculation is as follows:
\begin{align}
    m_i=\max \limits_{1<=j<=k} s_{ij}+e_{ij}
\end{align}
\begin{align}
    g_{ij}=\frac{s_{ij}+e_{ij}}{\lambda}-m_i
\end{align}
\begin{align}
    {sp}_{ij}=\frac{e^{g_{ij}}}{\sum_{j=1}^k e^{g_{ij}}}
\end{align}
When the scaling factor $\lambda$ is set to 1, ${sp}_{ij}$ is equal to $p_{ij}$. The value of $\lambda$ should be higher than 1 to make sure $g_{ij}$ is negative. To address overconfident issue, we set $\lambda$ to 1.3, which is acquired through several experiments.

So we take manual features with a total of $3+5k$ into consideration. 

\subsection{Representation learning features}
The other category is based on specific representations from models. We follow the pipeline but set the batch size to 1, so the output of trained model is relevant to the input example and the model states may imply information about selecting optimal answer. 

For each input example $i$ containing a question and a context, the pipeline will separate them with a special token, and generate the embedding and a sequence of hidden vectors from different hidden layers. The prediction is generated based on the final hidden layer. We denote the embedding as $v_i$, which is a fixed dimensional vector. And we denote the hidden states of model as a sequence of vectors $h_i=(h_{i,0},h_{i,1},...,h_{i,n})$, where $n$ is the number of layers \footnote{For RoBERTa-large, n is 24} and  $h_{i,m}$ is the corresponding hidden vector of $m$-th hidden layer. The vectors in $h_{i,m}$ have the same dimensionality as the embedding vector $v_i$, and we denote the dimensionality as $l$.

The large scale of $h_i$ may induce slow training and inference. So we only use the vector $h_{i,n}$ from last hidden layer and the average vector $A_i$ calculated as follows:
\begin{align}
    A_i=\frac{1}{n}\sum_{m=1}^n h_{i,m}
\end{align}
And we discover that adding embedding output $v_i$ is more effective, so we modify the calculation of $A_i$ to:
\begin{align}
    A_i=\frac{1}{n+1}\left(\sum_{m=1}^n h_{i,m}+v_i\right)
\end{align}

As a conclusion, we get three vectors $v_i$, $h_{i,n}$ and $A_i$ from the extractive MRC model. The three vectors have same dimensionality $l$, so we take representation learning features with a total of $3l$ into consideration.

\section{Experiments}
\subsection{Experiments settings}\label{section::4.1}
We take RoBERTa-large~\cite{liu2019roberta} provided in Hugging face transformers as our basic MRC model and use XGBoost~\cite{chen2016xgboost} provided by python library as the post-hoc calibrator.

We choose SQuAD 2.0 dataset~\cite{rajpurkar2018know} as our main dataset, and firstly fine-tune RoBERTa-large model on the dataset with two epochs as our basic model. And then we use the methods of adversarial examples generation and corresponding code provided in ~\cite{maharana2020adversarial} to generate adversarial examples. After obtained various adversarial examples with different amounts but all aggressive to the basic model, we use these data to verify the vulnerability of baseline model. Then considering the impact of the amount of training data on the results, we randomly separate 2k adversarial examples for calibrator training in the mixed setting, and use the rest as test sets for calibration results. Correspondingly, we separate half of SQuAD 2.0 dev set for calibrator training in both clean and mixed settings, and the rest for evaluation.

We use Natural Questions dataset~\cite{kwiatkowski2019natural} as a representative to evaluate the generalization performance. For convenience, we follow the setting of ~\cite{sen2020models} and use the provided scripts to convert Natural Questions datasets into a shared SQuAD 2.0 JSON format. We also use the same metrics as ~\cite{sen2020models} for better comparison with original SQuAD 2.0 dataset.

\subsection{Adversarial attack and generalization}

Followed ~\cite{maharana2020adversarial}, the methods of adversarial examples generation can be divided into two categories according to whether the language model is used in the process: negative for those are independent of language models and positive for the opposite. 

The negative category contains four methods: AddSentDiverse, AddKSentDiverse, AddAnswerPosition, and InvalidateAnswer. Part of these methods use templates or some heuristics to generate distracting sentences and then insert them randomly into context to disturb the model, and some apply deletion of crucial sentences to disturb the model. The positive category is composed of two methods: PerturbAnswer and PerturbQuestion. Both methods use language model to rephrase sentences into different forms with the same semantics. The detailed description and examples of these methods can refer to ~\cite{maharana2020adversarial}.

Considering that AddKSentDiverse has the same principle as AddSentDiverse but more aggressive, we ignore AddSentDiverse and only adopt AddKSentDiverse. PerturbAnswer is not suitable for our experimental scenario either, because our main dataset is SQuAD 2.0 that contains unanswerable questions. In summary, we apply four kinds of methods to generate adversarial examples: AddKSentDiverse, AddAnswerPosition, InvalidateAnswer, and PerturbQuestion. Tables \ref{tab:adversarial attack} shows datasets' sizes and the results of evaluating basic model on six datasets, where the model trained on SQuAD 2.0 merely chooses the answer with max softmax probability as output without using calibrator. According to ~\cite{maharana2020adversarial}, adding adversarial examples to train the basic model makes great improvement on adversarial datasets while degradation on original dataset.
\begin{table}
\centering
\begin{tabular}{lrrr}
\toprule
Testset            & size  & EM    & F1    \\
\midrule
SQuAD2.0-dev       & 11873  & 85.30  & 88.29 \\
\midrule
AddKSentDiverse    & 4586  & 49.96  & 53.41 \\
AddAnswerPosition  & 4355  & 64.50  & 68.72 \\
InvalidateAnswer   & 5861  & 65.96  & 65.96 \\ 
PerturbQuestion    & 3923  & 23.43  & 45.27 \\
\midrule
Natural Questions  & 3369  & 45.89  & 53.30 \\
\bottomrule
\end{tabular}
\caption{Baseline results without using calibrator on six datasets. PerturbQuestion is the most aggressive, resulting in the most decline.}
\label{tab:adversarial attack}
\end{table}

\begin{table*}
\centering
\begin{tabular}{c|l|rrr|rrr}
\toprule
 \multicolumn{2}{c|}{Trained on clean data}& \multicolumn{3}{c|}{AddKSentDiverse}  & \multicolumn{3}{c}{SQuAD 2.0 dev}   \\

Feature kind& Feature selection  & Acc  & EM  & F1 & Acc & EM & F1 \\
\midrule
\multicolumn{2}{c|}{Baseline(without calibrator)}    & 55.68  & 50.50  & 54.12 & 86.39  & 84.10 & 87.39\\
\midrule
manual&$c_i+q_i+l_{i0}$   & 55.68  & 50.58  & 54.12 & 85.90  & 83.91 & 87.29\\
&$+p_{ij}$       & 55.57  & 50.66  & 54.3 & 86.02 & 84.00  & 87.25 \\ 
&$+p_{ij}+E_i$    & 55.80  & 50.81  & 54.38 & 86.10 & 84.02  & 87.30  \\
&$+sp_{ij}$   & \textbf{55.99}  & \textbf{51.40}  & \textbf{54.62} & 85.80 & 83.90  & 87.20  \\
&$+sp_{ij}+E_i$    & 55.57  & 50.81  & 54.46 & 85.87 & 83.88  & 87.21  \\
\midrule
representation&$+v_i$   & 55.72  & 50.54  & 54.13 & 85.31 & 83.43  & 86.97  \\
learning&$+h_{i,n}$  & 55.38  & 50.35  & 54.00 & \textbf{86.41} & \textbf{84.12}  & \textbf{87.38}  \\
&$+A_i$  & 55.76  & 50.62  & 54.14 & 86.31 & 84.02  & 87.34  \\
\bottomrule
\end{tabular}
\caption{
The results on AddKSentDiverse when calibrator only trained on clean original data. All features have been described in section \ref{section::3}. Baseline result is the output of basic model without calibration. Applying manual features to train the calibrator can improve the performance on AddKSentDiverse. Representation learning features just maintain the baseline. Applying the mixture of manual features and representation features has similar results with only apply manual features to train, which we omit in the results.}
\label{tab:clean data AddKSentDiverse}
\end{table*}

\begin{table*}[!h]
\centering
\begin{tabular}{c|l|rrr|rrr}
\toprule
 \multicolumn{2}{c|}{Trained on mixed data} & \multicolumn{3}{c|}{AddKSentDiverse}  & \multicolumn{3}{c}{SQuAD 2.0 dev}   \\

Feature kind & Feature selection      & Acc  & EM  & F1 & Acc & EM & F1 \\
\midrule
\multicolumn{2}{c|}{Baseline(without calibrator)}           & 55.68  & 50.50  & 54.12 & 86.39  & 84.10 & 87.39\\
\midrule
manual &$c_i+q_i+l_{i0}$   & 56.38  & 51.28  & 54.93 & 85.99  & 83.83 & 87.17\\
&$+p_{ij}$          & 61.60  & 57.04  & 60.91 & 85.04  & 83.00 & 86.26 \\ 
&$+p_{ij}+E_i$      & 61.64  & 57.08  & 60.94 & 84.89  & 82.94 & 86.25  \\
&$+sp_{ij}$         & 61.87  & 57.42  & 61.36 & 85.18  & 83.11 & 86.43  \\
&$+sp_{ij}+E_i$     & 61.64  & 57.12  & 60.98 & 85.06  & 82.94  & 86.28  \\
\midrule
representation&$+v_i$             & 57.46  & 52.47  & 56.27 & 85.31  & 83.19  & 86.61  \\
learning&$+h_{i,n}$         & 63.77  & 59.59  & 64.17 & 86.10  & 83.81  & 87.11  \\
&$+A_i$             & 63.81  & 59.74  & 64.59 & 86.29  & 84.00  & 87.27  \\
\midrule
manual+&$+v_i+sp_{ij}$     & 62.14  & 57.54  & 61.45 & 85.36  & 83.29  & 86.63  \\
representation&$+h_{i,n}+sp_{ij}$ & 65.20  & 60.94  & 65.42 & 86.15  & 83.90  & 87.11  \\
learning&$+A_i+sp{ij}$      & \textbf{67.25}  & \textbf{62.99}  & \textbf{67.76} & \textbf{86.39}  & \textbf{84.12}  & \textbf{87.41}  \\
\bottomrule
\end{tabular}
\caption{
The results on AddKSentDiverse when calibrator is trained on the mixture of original data and 2k AddKSentDiverse data. The description of features and baseline is the same as table \ref{tab:clean data AddKSentDiverse}.}
\label{tab:mixed data AddKSentDiverse}
\end{table*}

\subsection{Calibrator}\label{section::4.3}
A good calibrator should improve the performance on adversarial and generalization dataset, and maintain even improve the performance on the original dataset. We use data described in section \ref{section::4.1} to train and evaluate the calibrator. The calibrator is trained in two settings: clean data and mixed data.

\subsubsection{Clean data}
Under this setting, the calibrator is only trained on the separated SQuAD 2.0 dataset. We suppose that if qualified features are extracted, the calibrator can improve the performance on the distribution-shift dataset even trained on the original dataset.

We take manual features and representation learning features described in section \ref{section::3} into consideration. Accuracy of calibrator, EM and F1-score are the metrics to be evaluated. We take AddKSentDiverse as a representative to demonstrate varying results under different selection of features in table \ref{tab:clean data AddKSentDiverse}.

From the experimental results, we found that manual features can be helpful when calibrator only trained on clean data. It can improve performance of adversarial dataset by 1\% while degradation by less than 0.2\% on the original dataset. Since the calibrator is ignorant of distribution-shift data, it can't utilize representation learning features and just maintain the baseline result. Among manual features, $E_i$ and $sp_{ij}$ we proposed can be most informative to calibration. It seems that improving the performance of distribution-shift data without sacrificing the original performance is infeasible when calibrator is only trained on the clean data. Further exploration on better features is required.

\subsubsection{Mixed data}
We also train the calibrator on the mixture of original data and 2k target test dataset, where target dataset is one of adversarial or generalization datasets. ~\cite{maharana2020adversarial} used the mixed data to train the QA model, which leads to degradation of original dataset due to the shift of data distribution. We suppose the calibrator can alleviate this issue.

The consideration of feature selection and evaluation metrics is the same as that of clean data setting. We also take AddKSentDiverse as a representative to demonstrate varying results under different selection of features in table \ref{tab:mixed data AddKSentDiverse}.

Table \ref{tab:mixed data AddKSentDiverse} shows that the access to target examples can bring great improvement on target testset. When only exploring manual features, the performance on the target testset can be improved by 6\% to 7\% on all metrics while degradation on original dataset by about 1\%. Representation learning features can be great helpful not only to improve the target performance by 9\% but also to keep original performance drop less than 0.2\%. The combination of manual and representation learning features can improve the target performance by nearly 13\%, and improve the performance of original dataset.

\subsubsection{Comparison on different mixed methods}

Due to the limitation of paper length, we can't list results of all feature combinations on all test datasets, which will be available in our repository. But to prove that the effect of the calibrator is not limited to one particular dataset, we list the results of best features selection on all test datasets in table \ref{tab:data all}. The results of clean data and single mixed correspond to the above descriptions. In single mixed, the result of SQuAD 2.0 dev is the average across various mixed data. To better prove the improvement of robustness, we extract 1k from each distribution-shift data and mix them with original data to train the calibrator, and list results in table \ref{tab:data all}. From results, our calibrator can make effects whether trained on single or all mixed data. For the latter case, the calibrator can improve the generalization ability with little generalization data.

\section{Analysis}\label{section::5}

\subsection{Analysis of the bad cases of baseline}

\begin{table*}
\centering
\begin{tabular}{l|rr|rr|rr}
\toprule
Test data       & \multicolumn{2}{c|}{Clean data} & \multicolumn{2}{c|}{Single mixed}&
\multicolumn{2}{c}{All mixed} \\
              & \multicolumn{1}{c}{EM}    & \multicolumn{1}{c|}{F1}
              & \multicolumn{1}{c}{EM}    & \multicolumn{1}{c|}{F1}
              & \multicolumn{1}{c}{EM}    & \multicolumn{1}{c}{F1} \\
\midrule

AddKSentDiverse      & 50.50+0.9  & 54.12+0.5 & 50.50+12.49  & 54.12+13.64 & 50.50+10.25  & 54.12+11.18\\
AddAnswerPosition    & 65.01+0.51  & 69.49+0.35 & 65.01+11.91  & 69.49+12.36 & 65.01+5.69  & 69.49+5.50\\
InvalidateAnswer     & 64.98+6.03  & 64.98+6.03 & 64.98+10.7  & 64.98+10.7 & 64.98+1.01  & 64.98+1.01\\ 
PerturbQuestion      & 23.56+0.72  & 45.81+0.51 & 23.56+14.25  & 45.81+8.86   & 23.56+9.51  & 45.81+6.28\\
\midrule
Natural Questions    & 44.27+0.1  & 51.96+0.1 & 44.27+1.5  & 51.96+2.0 & 44.27+1.38  & 51.96+1.44\\
SQuAD 2.0 dev        & 84.10+0.02 & 87.39-0.01 & 84.10+0.1 & 87.39+0.05 & 84.10+0.1 & 87.39+0.1\\
\bottomrule
\end{tabular}
\caption{The best results on all datasets. Clean, single and all mixed refer to cases when calibrator is trained on data described in section \ref{section::4.3}}
\label{tab:data all}
\end{table*}

In order to figure out why the performance of fine-tuned model dropped dramatically when applying adversarial or generalization examples, we analyzed the bad cases based on table \ref{tab:adversarial attack}. We defined any example whose final prediction has lower F1-score than average as a bad case. Then we explored the top $k$ candidates provided by the model corresponding to this bad case, calculated the F1-score respectively, and labeled the best of top $k$ candidates. If the answer with max softmax probability is not the best, it means there are better candidates in top$k$ predictions. We first made statistics on the number of bad cases in all datasets and proportion of examples with better candidates. We found that almost 90\% of bad cases can find a better candidate among top $k$ predictions. We also make this analysis on all examples of the whole datasets rather than limited to bad cases. We found that larger proportion of examples with better candidates in adversarial and generalization dataset comparing to only less than 15\% of original dataset. The result is presented in table \ref{tab:bettersize}.
\begin{table}
\centering
\begin{tabular}{lrr}
\toprule
Testset            & size  & better-size  \\
\midrule
SQuAD2.0-dev       & 11873  & 1778   \\
AddKSentDiverse    & 4586  & 2062   \\
AddAnswerPosition  & 4355  & 1536  \\
InvalidateAnswer   & 5861  & 1995  \\ 
PerturbQuestion    & 3923  & 2757  \\
Natural Questions  & 3369  & 1747  \\
\bottomrule
\end{tabular}
\caption{The result on the number of examples with better candidates among top $k$ candidates on all datasets. }
\label{tab:bettersize}
\end{table}

\begin{figure}
    \centering
    \includegraphics[width=.85\linewidth]{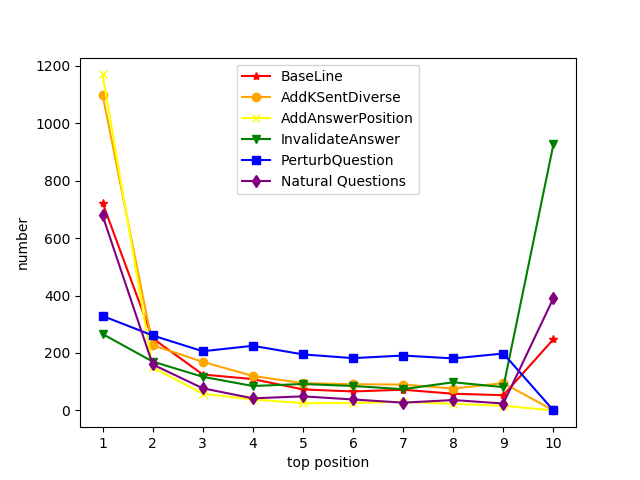}
    \caption{The label of best answers among top $k+1$ candidates. We must emphasis that top 0 means the answer with max softmax probability instead of top 1. }
    \label{fig:my_fig}
\end{figure}

So we came to the conclusion that the shift of data distribution makes the ranking based on softmax probability of baseline model unreliable. We used the labels of best among top $k$ candidates to draw a line chart to show the shift in alignment between examples of high confidence and empirical likelihoods, which is presented in figure \ref{fig:my_fig}. From the graph, we found that most of best answers is limited to top 3 answers, which means the shift of data distribution didn't cause huge deviation on the ranking. So the calibrator we proposed can make great improvement without sacrificing original performance. But for InvalidateAnswer and Natural Questions datasets, examples with better candidates focus on the second half are more than the original baseline, so the improvement characteristics of the calibrator is different from the other datasets. InvalidateAnswer depends more on manual features, especially text length of predictions, than others due to its special construction. Natural Questions dataset needs external knowledge to make more improvement.

\subsection{Analysis of features selection}
The selection of features is crucial to the improvement of calibrator no matter which dataset. From table \ref{tab:clean data AddKSentDiverse} and table \ref{tab:mixed data AddKSentDiverse} and results on other datasets which will be available in our repository, we find that manual features improve target performance with slight degradation on original dataset, while representation learning features perform better on original dataset and can make great improvement on target dataset under mixed data setting. $sp_{ij}$, $E_i$ and $A_i$ can be most helpful among respective category. $l_ij$ has no effect on most of datasets under clean data setting, but is very useful under mixed data setting.

When multiple features are selected, the order of different features will have a certain impact on the results, but the impact is not as big as the selection of features. So results we reported are the average of different orders. More kinds of features and their combination need further exploration.

\section{Conclusion}

We use the calibrator as a reranker to improve performance of adversarial and generalization dataset without sacrificing the original performance. We take manual features and representation learning features into consideration. When the calibrator is only trained on the clean data, the adversarial performance can improve by 1\% while degradation by less than 0.2\% on original dataset. When the calibrator is trained on the mixture of original and adversarial data, the target performance can improve by more than 10\% while maintaining the original performance. And our calibrator only takes about ten minutes to train and is very easy to use as a post-hoc structure behind MRC model. We also demonstrate the impact of distribution-shift data on model. To summarize, our calibrator is simple, effective, and has potential to be practical application and extended to other NLP tasks.

\bibliographystyle{named}
\bibliography{reference}

\end{document}